\documentclass[
]{ceurart}

\pdfoutput=1
\usepackage{amsmath} 
\usepackage[font=small,skip=0pt]{caption}
\usepackage[noorphans,vskip=0ex,indentfirst=false]{quoting}

\begin{document}

\copyrightyear{2022}
\copyrightclause{Copyright for this paper by its authors.
  Use permitted under Creative Commons License Attribution 4.0
  International (CC BY 4.0).}

\conference{In: R. Campos, A. Jorge, A. Jatowt, S. Bhatia, M. Litvak (eds.): Proceedings of the Text2Story'22 Workshop, Stavanger (Norway), 10-April-2022}

\title{Extracting Impact Model Narratives from Social Services' Text}

\author[1]{Bart Gajderowicz}[%
orcid=0000-0001-6201-8781,
email=bartg@mie.utoronto.ca,
url=http://bartg.org
]
\author[1]{Daniela Rosu}[%
orcid=0000-0002-5877-9681 ,
email=drosu@mie.utoronto.ca,
]
\author[1]{Mark S. Fox}[%
orcid=0000-0001-7444-6310,
email=msf@mie.utoronto.ca,
url=http://eil.utoronto.ca,
]

\address[1]{Department of Mechanical \& Industrial Engineering, University of Toronto, 5 King's College Road, Toronto, Ontario, M5S 3G8, Canada}

\begin{abstract}
Named entity recognition (NER) is an important task in narration extraction. Narration, as a system of stories, provides insights into how events and characters in the stories develop over time. This paper proposes an architecture for NER on a corpus about social purpose organizations. This is the first NER task specifically targeted at social service entities. We show how this approach can be used for the sequencing of services and impacted clients with information extracted from unstructured text. The methodology outlines steps for extracting ontological representation of entities such as needs and satisfiers and generating hypotheses to answer queries about impact models defined by social purpose organizations. We evaluate the model on a corpus of social service descriptions with empirically calculated score.
\end{abstract}


\maketitle

\section{Introduction}
In Canada, there are over 65,000 social service providers, each delivering a specialized set of services administered as a variety of programs to different populations in need. These populations include the approximately 4.9 million Canadians living in poverty, along with the additional 401,000 new immigrants and 44,000 refugees that immigrated to Canada in 2021 alone. The Canadian federal government plays an key role in coordinating with private and public agencies to fund the vast majority of social programs active today. However, the task of service provisioning and management is largely the responsibility of individual social purpose organizations (SPO). Navigating the myriad of SPOs to address an individual's (client) needs is a daunting task fraught with errors. Determining the efficacy of the services assigned to a client is also a challenge. As of this writing, no system can manage such a large network of services and clients in a cohesive manner, but such a system would benefit all stakeholders involved. 

In our previous work we have developed an ontology that can be used by an SPO to represent its Impact Management models~\cite{CIDS2021}. The term impact ``\textit{refers to the intended and unintended (positive or otherwise) changes (outputs, outcomes) that occur across the organization (within and/or across its programs) and with its stakeholders (including users, clients, partners, etc.) over a period of time (short term, long term) as a result of the organization's activities.}''\footnote{\url{https://innoweave.ca/en/modules/impact-measurement}} Experts have developed numerous Impact Models to help SPOs articulate the change they seek to achieve and how that change is achieved~\cite{PCI1979,Weiss1997,Earl2001, Harries2014, Nicholls2012}. The Common Impact Data Standard (CIDS)~\cite{CIDS2021} ontology defines classes and relationships that span impact modelling concepts such as Program, Service, Activity, Stakeholder, Outcome, Indicators and Risk. It can be used to define the services an SPO provides and the requirements needed for a client to receive a service. Our most recent ontology research extends CIDS to include client needs (e.g., housing, food) and various ways they can be satisfied (e.g. women shelters, food banks).

In this paper, we describe our efforts in addressing the problem of matching client needs to SPO services. In order to match client needs, we must represent the services an SPO provides and how they satisfy needs, and the characteristics (e.g., age, gender, occupation) and needs of clients for whom they were designed. Although we have the means to represent an SPOs impact model, the information needed to instantiate each SPO's model is buried in a variety of textual sources, such as service descriptions, client success stories, and eligibility criteria. 

This paper presents an approach, based on Named Entity Recognition (NER), to extracting an SPO's impact model, i.e., ``narrative'', from various text sources.  NER is a crucial component in understanding the narrative of a given text. By narrative, we mean a ``\textit{system of stories structured in such a way as to achieve a rhetorical purpose or vision}''~\cite{Ruston2016}.  In the context of SPOs, we define each service as a ``story'' describing what they offer, to whom, how, and when. The narrative, then, is a system of services that guide clients through various programs towards achieving their goals. The extracted information can then be provided in the same language as the problem domain, allowing for culture-wide, community-based, and individual-level analysis~\cite{Ruston2016}.

There are several \textbf{challenges} that we face when extracting terms that represent client needs and specific resources that services provide. Consider the example
\begin{quoting}
    ``\textit{We welcome clients of all ages and offer services that benefit our core clients (young families, chronically homeless): mental health, education needs; community outreach.}''
\end{quoting}
that provides required information but is hard to parse. Firstly, there is a lack of language models for identifying social service client characteristics and needs, and how their services satisfy needs. Often, vague descriptors cause confusion about entities: what is a program, what is a service, what is the resource, what is the need. There are no standardized labels across SPO programs, services, clients, and eligibility criteria. Structured data about services is limited, while unstructured text describes various aspects of the service that are hard to infer, such as promoting services versus listing service details, or describing serviced communities versus listing client requirements. Finally, information related to the scheduling or expected sequence of services is often incomplete or unknown, such as quality, service capacity, or availability.

\section{Related Work}
\label{sec:related}
\subsection{Named Entity Recognition}
\label{subsec:ner}
Named Entity Recognition (NER) is a method for identifying the types of terms in unstructured text. Common terms include a person, organization, place, date, currency, and numbers~\cite{Weischedel2013}. As will be described in the following sections, three main approaches used are: 1) a rule-based models for identifying types of terms, 2) a learned model that can infer types found in a training set of text and types, and 3) a hybrid rule-learned model. The rule-based method can identify patterns in text when proper sentence grammar is not followed, or entities are not common enough to be found in a training set (e.g. business names). However, a rule-based method requires manual evaluation of the data and manual rules construction for observed patterns. The trained method can match words in a sentence to its learned vocabulary and assign their type, but is limited to words in the training set. Hybrid models try to take advantage of both rules and learned methods, making best guesses to infer entities not present in the training set. 

There is a number of pre-trained language models capable of named entity recognition. Each one is trained on either a specialized or general dataset. Schmitt et al.~\cite{Schmitt2019} performed a comprehensive analysis of the performance and applicability of the five most popular packages, including StanfordNLP, NLTK, OpenNLP, SpaCy, and Gate with new ones being developed on an ongoing basis, such as HuggingFace~\cite{Wolf2020}. These packages are trained on varying datasets, the two largest being Common Crawl (\url{http://commoncrawl.org}), a database of content crawled on the internet, and English Wikipedia (\url{https://www.wikipedia.org}). NER benchmark datasets include MUC-6~\cite{Grishman1995} and MUC-7~\cite{Chinchor1997} and ACE~\cite{Doddington2004}. The resulting models generally provide support for a standard list of entity types~\cite{Weischedel2013}. Unfortunately, existing models are not well suited for social services as they have not been trained on related corpa identify required entities.

\subsection{Other Methods}
\label{subsec:othermethods}
In this section, we highlight several approaches and their methods that identify entities in the text and can assist in building an SPO's impact model narrative. \textbf{Linguistic properties} alone provide a great deal of structure to the text being analyzed. For example, Chiarello et al.~\cite{Chiarello2018} use linguistic features to identify stakeholders across documents, while Hussain et al.~\cite{Hussain2021} rely on grammar rules to generate narratives, identify keywords and extract important phrases in social media posts. \textbf{Query-driven} methods rely on a ``seed'' query and external vocabulary to guide the search algorithm and find a suitable label in order to  perform query answering~\cite{Sciore2015ch6}, query modelling~\cite{Balog2010,Craswell2009}, and query extensions~\cite{Balog2011} tasks. \textbf{Rule-Based methods} are suitable when a training corpus does not exist, and a set of a \textit{priori} rules provide context for extracting various entities. This includes predefined rules for finding clues for query answering~\cite{Garigliotti2017} and grammar rules for narrative extraction~\cite{Hussain2021,Quaresma2020}, and to reason about extracted entities~\cite{McCord2012}.

\textbf{Statistical models} rely on data-driven algorithms and encompass both frequency-based and probability-based models~\cite{Hong2010, Robertson2009} for ranking found entities~\cite{Oza2021}, group generalization~\cite{Balog2010,Balog2011}, and calculating similarity scores between documents~\cite{Garigliotti2017}.  \textbf{Machine learning} methods include deep learning architectures for NER tasks~\cite{Li2022}. Several have characteristics useful for narrative extraction, such as temporal factors, rules, or linguistic properties, and utilize methods such as BiLSTMs~\cite{Taille2020,Affi2021,Lample2016,Jie2020}, ELMo~\cite{Affi2021,Peng2019,Ulcar2021,Dogan2019}, and BERT~\cite{Moon2019,Souza2019,Liang2020,Zhou2021,Rottger2021,Vani2020}.

\section{Methodology}
\label{sec:methodology}
\begin{figure}
  \centering
  \includegraphics[width=0.9\linewidth]{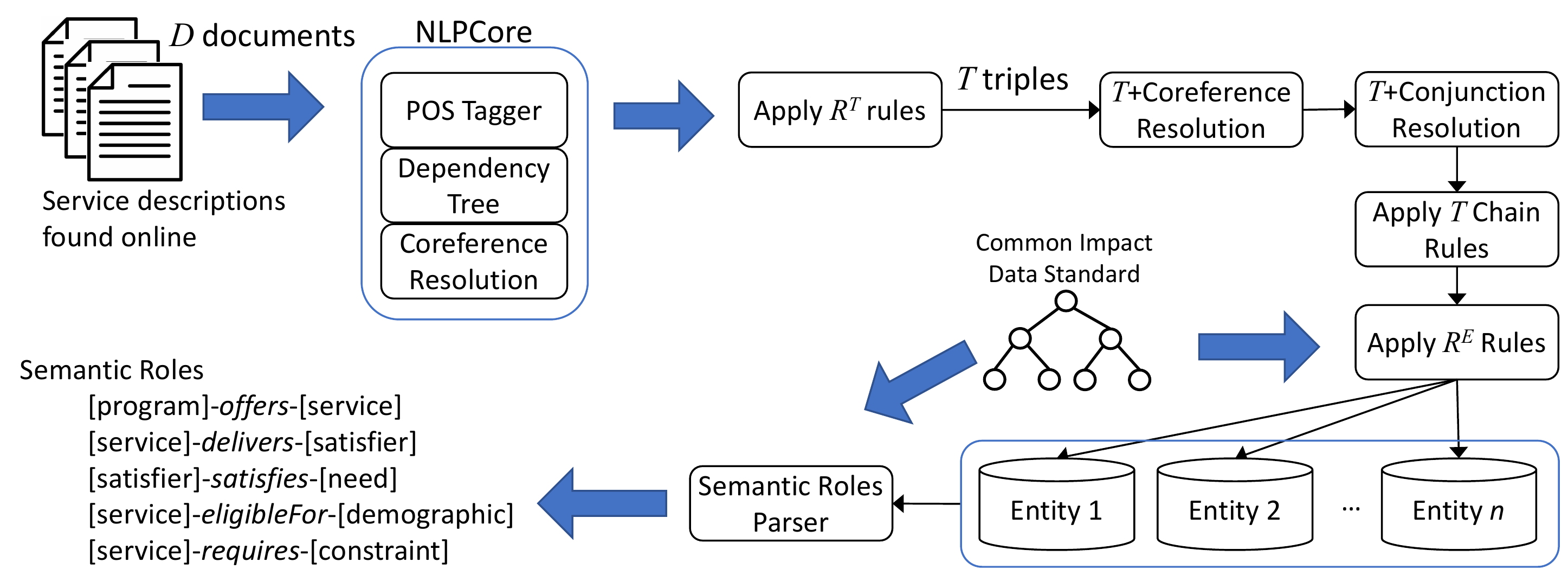}
  \caption{Named Entity Extraction Architecture.}
    \label{fig:arch}
\end{figure}

This section summarizes our methodology for performing the NER task in the SPO domain. Our proposed NER architecture is depicted in  Figure \ref{fig:arch}. The input is a corpus of unstructured text describing SPOs. It incorporates the Common Impact Data Standard~\cite{CIDS2021} ontology to identify which entities to extract, then again to generate semantic roles by providing the semantic relationships between the entities. The terms we are interested in are listed in Table \ref{tab:cids_types}. They capture key concepts in describing an SPOs logic model, and form the basis of their ``narrative'' in how services are delivered to clients, what needs they are satisfying, how, and when. We also introduce several definitions used by our NER model in Table \ref{tab:defs}, and related equations.
\begin{table*}
\centering
\footnotesize
\setlength\extrarowheight{-4pt}
\setlength\tabcolsep{3pt}
\caption{SPO-related entity types.}
\label{tab:cids_types}
\begin{tabular}{p{100pt}p{300pt}}
\toprule
\textbf{Entity Type}         & \textbf{Description}                          \\
\midrule
Program   Name               & Name of the   program.                        \\
Need   Satisfier             & Label for   the need satisfier that focuses more on the offered resources.      \\
Client   Characteristic      & A client's   characteristics, as defined by the service provider.           \\
Need                         & The   client's need being addressed.                            \\
Desired   State (outcome)    & The desired state of the client, which may more broadly describe the outcome.                        \\
Service   Description        & A broader description of the service being offered.  \\
Client Description           & A broader description of the client. \\
Need   Satisfier Description & A broader description of the need satisfier. \\
Required   Criteria          & Criteria for using a service as representing a requirement or eligibility.     \\
\bottomrule                                                                             
\end{tabular}
\end{table*}
\begin{table*}
\centering
\footnotesize
\setlength\extrarowheight{-4pt}
\setlength\tabcolsep{3pt}
\caption{Definitions.}
\label{tab:defs}
\begin{tabular}{c p{330pt}}
\toprule
\textbf{Variables}         & \textbf{Description}                          \\
\midrule
$d\in D$   & Documents $d$ in a corpus $D$.  \\
$e\in E$    &   Entities to be extracted from text in $D$.\\
$t_x\in T$  &   Subject-predicate-object triples.\\
$s,p,o$     &   Each triple is comprised of three slots, where $t_x = {s(subject), p(predicate), o(object)}$\\
$T_d \subseteq T$ & A subset of triples in $T$ extracted from document $d$. \\
$r^t_i \in R^T$ &   Rules identifying triples in $t_x\in T$.\\
$r^E_i \in R^E$ &   Rules identifying entities in $e\in E$.\\
$mcc_i$     &   Matthews correlation coefficient (MCC) score for rule $r_i^E$.\\
$r^E_i (T_d)$   &   Classification of entity e by $r_i^E$ from triples in $T_d$.\\
$\widetilde{w}_i$   &   Weight of rule $r_i^E$ in correctly identifying an entity $e$, as per Equation \ref{eq:wi}.\\
\bottomrule                                                                             
\end{tabular}
\end{table*}
\begin{equation}
    r^E_i(T_d) = 
     \begin{cases}
       \text{1,} &\quad\text{if }mcc_i >0 \\[-4pt]
       \text{0,} &\quad\text{otherwise}
     \end{cases}
     \label{eq:rei}
\end{equation}
\begin{equation}
    \widetilde{w}_i = r^E_i(T_d) \times mcc_i
    \label{eq:wi}
\end{equation}
\begin{equation}
    w_e = \dfrac{\sum \widetilde{w}_i}{|r^E|} \text{ for all rules }r^E_i\in r^E \text{ that extract entity }e.
    \label{eq:we}
\end{equation}
\subsection{Annotating with Linguistic Properties}
Our method begins by relying on a Stanford NLPCore parser~\cite{Manning2014} to generate a set of linguistic properties about the service descriptions. We use its part-of-speech (POS) tagger to identify nouns, verbs, adjectives, and so on. Second, the parser creates a dependency tree identifying word modifiers, conjunctions, as well as subjects, predicates, and objects. Third, the parser generates coreference resolutions between terms, associating pronouns like ``they'' and ``our'' with the nouns or proper nouns they refer to. Accuracy and further processing is limited by the accuracy of the dependency trees and coreference resolutions generated by the NLPCore parser. 

\subsection{Semantic Role Triple Extraction}
Once the parser has annotated the text with linguistic properties, custom rules $r_x^T\in R^T$ combine key dependencies to form subject-predicate-object triples $t_x\in T$. In some literature, the triple relation is referred to as subject-verb-object (SVO), but $T$-triples represent a broader structure that does not rely on verbs as predicates alone. Each triple contains three slots: a subject (s), a predicate (p), and an object (o), forming the structure:\\
    \centerline{$t_x$ = \{\;s(``subject''), \;p(``predicate''), \;o(``object'')\;\}}.
For example, consider the sentence ``St. Mary's provides education services.'' Here we see that ``St. Mary's'' is the subject, ``provides'' is the predicate, and ``education services'' is the object. Consider a rule $r^T_i$ where, given the three terms $A$, $B$, and $C$, and dependencies $(nsubj, obj, obl)$\\
\vspace{-15pt}
\begin{center}
\setlength\extrarowheight{-3pt}
\setlength\tabcolsep{3pt}
\begin{tabular}{ll}
If &  a $nsubj$ dependency exists between $B$ and $A$,\\
  &  an $obj$ dependency exists between $B$ and $C$,\\
  &  an $obl$ dependency does not exist between $B$ any other term, \\
Then   & $t_x = \{s(A), p(B), o(C)\}$.
\end{tabular}
\end{center}
\vspace{-2pt}
\noindent
By applying this rule to the sentence above, we can infer the $T$ triple:\\
    \centerline{$t_x$ = \{\;s(``St Mary's''),\; p(provides), \;o(``education services'')\;\}.}
While this example rule is easily inferred from the dependencies alone, 18 rules $r^T_i \in R^T$  have been empirically identified to extract subject-predicate-object relationships. 

\subsection{Coreference and Conjunction Resolution}
Next, each $T$-slot is extended with their coreference and conjunction terms, if any, using a depth-first search. For example, in the sentences ``\textit{St. Mary's provides education services. They also prepare hot meals.}'', the pronoun ``They'' refers to the proper noun ``St. Mary's''. Hence we infer that in addition to ``education services'', ``St Mary's'' also provides ``hot meals'', giving:
    \centerline{$t_1$ = \{\;s(``St Mary's''), \;p(``provides''),\; o(``education services'')\;\}}\\
    \centerline{$t_2$ = \{\;s(``St Mary's''), \;p(``prepares''),\; o(``hot meals'')\;\}}
    
Next, we resolve conjunctions with terms in each $T$-slot. Conjunctions are lists of terms connected by terms like a comma, ``and'' and ``or''. For example, given the sentence ``\textit{St. Mary's provides education services, a soup kitchen, and religious counselling.}'',
we see that all terms following ``provides'' are of the same type, a ``need satisfier.''  

Like subjects and objects, the predicate can also be a conjunction. For example, in the sentence ``\textit{St. Mary's provides education services and prepares hot meals.}'', ``provides'' and ``prepares'' are both verbs connected as a conjunction in the dependency tree, and hence both have ``St Mary's'' as their subject. However, they each have their own object, producing two $T$-triples, namely\\
    \centerline{$t_1$ = \{\;s(``St Mary's''),\; p(``provides''), \;o(``education services'')\;\}}\\
    \centerline{$t_2$ = \{\;s(``St Mary's''),\; p(``prepares''), \;o(``hot meals'')\;\}}
To ensure we capture all combinations of $T$-triples, our model uses a depth-first search to generate hypotheses for all combinations of connected subjects, predicates, and objects.

\subsection{Chaining Rules: From Triples To Stories}
Given a list of $T$-triples, and coreferences and conjunctions resolved, we build a chain of $T$-triples that provide additional structure to the terms in the text. The rules simply connect object slot $(X)$ values in one $t_1$ triple to subject slot $s(X)$ values in another $t_2$ triple, 
\vspace{-2pt}
\begin{center}
\setlength\extrarowheight{-3pt}
\setlength\tabcolsep{3pt}
\begin{tabular}{ll}
If &	  $t_1 = \{s(A),p(B),o(C)\}$ and $t_2 = \{s(C),p(D),o(E)\}$        \\
Then &   $[\{s(A),p(B),o(C)\}; \{s(C), p(D),o(E)\}]$ is a $T$-chain and\\
    & $t_3=\{s(A), p(D),o(E)\}$.
\end{tabular}
\end{center}
\vspace{-2pt}
In the sentence ``\textit{St. Mary's provides education services to adult learners.}'' we see two $T$-triples:\\
    \centerline{$t_1$ = \{\;s(``St. Mary's''),\;p(``provides''),\; o(``education services'')\;\}}\\
    \centerline{$t_2$ = \{\;s(``education services''),\; p(``to''),\; o(``adult learners'')\;\}}
Chaining them together with the rule above using the ``education services'' term, we can infer that ``St Mary's'' offers services to adult learners, generating a new $T$ triple:\\
    \centerline{$t_3$ = \{\;s(``St. Mary's''), \;p(``to''),\; o(``adult learners'')\;\}}
\subsection{Named Entity Extraction Rules}
From $T$-triples and $T$-chains, we can apply additional rules $r_i^E \in R^E$ to extract named entity types. For example, we see that ``St Mary's'' is the program, ``education services'' is the need satisfier, and ``adult learners'' are the clients. The rules utilize all available information about the text, including POS tags, dependencies, and their $T$-slots. For example, given terms $A$, $B$, $C$:
\vspace{-2pt}
\begin{center}
\setlength\extrarowheight{-3pt}
\setlength\tabcolsep{3pt}
\begin{tabular}{ll}
    If    &   $t_x = \{s(A), p(B), o(C)\}$, where $A$ is a proper noun, $B$ is a synonym for ``offers'',      \\
          &   $B$ is a 3rd person singular present verb, and $C$ is a plural noun                   \\
    Then  & $A$ is a program and $C$ is a need satisfier.
\end{tabular}
\end{center}
\vspace{-2pt}
Here, synonyms for ``offers'' have been empirically identified as keywords used by services providers to describe what need satisfiers they offer to clients, and include the terms "provides", "offers", "offer", "provide", "provided", "offered", and "offering". Similar extractions can be performed for additional semantics defined by Common Impact Data Standard~\cite{CIDS2021} such as, [program]-offers-[service description], [service description]-delivers-[need satisfier]. [need satisfier]-satisfies-[need]. [service description]-eligibleFor-[client demographic], and [service description]-requires-[constraint].
\section{Evaluation}
The evaluation of our model is based on the performance of each rule, aggregated by entity type into a single entity score, namely $w_e$. The data contains information about SPOs, provided by Help Seeker Technologies (\url{https://helpseeker.co}). The testing data consists of 16,048 documents $d\in D$ that contain SPO descriptions. Of those, 7,359 documents had a total of 76,592 unique $T$-triples extracted. Constructed from the triples, there were 147,299 $T$-chains found in 6,260 documents. Of those documents that had $T$-triples, 4,860 descriptions had at least one term extracted. In total 366,588 terms were extracted, and of those 48,729 were assigned an entity type using rules in $R^E$.

To evaluate the model, a number of documents were selected randomly for each entity, and the extracted terms were manually analyzed. Table \ref{tab:rules} lists the results of each rule $r_i^E$ identifying an entity $e$. The rule's label indicates its number and which slot was used (e.g. Rule = ``o-45'' means the ``o'' slot for rule 45). All rules rely on $T$-triples. Those marked with ($\sharp$) also rely on $T$-chains. Each entity $e$ extracted from a document's triples $T_d$ was classified as correct (1) or incorrect (0), as per Equation \ref{eq:rei}.

We note that not all entities have the same number of rules and not all entities are covered equally. For example, the Need Satisfier entity has the largest coverage with 13 rules while Client Characteristics, Desired States, and Required Criteria only have one. We also note that the MCC score is sensitive to large discrepancies between true (TP,TN) and false (FP,FN) values. Rules that identify significantly more true values but are not good at excluding false values can produce a negative MCC score, despite a high F-score, as marked by (\S), and include Client Characteristic and Required Criteria. 

The model's performance is evaluated by its aggregate score for each entity, namely $w_e$, as per Equation \ref{eq:we}. The evaluation uses the ROC-AUC score to determine whether a high $w_e$ weight correlates with correct classification. The results are listed in the AUC column of Table \ref{tab:rules}. Any $AUC\geq 0.7$ is considered acceptable, and marked by ($\flat$). 

Based on the $w_e$ score, the model has good performance on extracting the ``Client Description'', ``Need Satisfier'', ``Need Satisfier Description'' and ``Program Name'' entities. The model also performs well with a high F-score on the ``Client Characteristic'' and ``Requirement Criteria'' entities but resulted in a low $w_e$ due to a negative MCC score. In cases where the MCC score was negative with a high F-score, we point out that if accuracy metrics (precision and recall) for the rule are high, the rule performed well on NER tasks but is limited to true positives only.

Relying on the entities that were correctly extracted, we can construct a set of semantic roles to build SPO narratives. For example, consider a particular program that delivers language classes (need satisfier) to new immigrants (a client characteristic). We can specify what requirements these clients must meet before receiving these satisfiers, such as language skill assessments. Knowing that another program offers language skill assessments, we can connect the two programs, defining a chain of SPO programs. 

\begin{table*}
\centering
\footnotesize
\caption{Rule statistics and score based on MCC, and an aggregate model score $w_e$ evaluated by AUC.}
\label{tab:rules}
\setlength\extrarowheight{-4pt}
\setlength\tabcolsep{3pt}
\begin{tabular}{p{3pt}p{20pt}p{16pt}p{16pt}p{16pt}p{16pt}p{17pt}p{30pt}p{16pt}|p{3pt}p{20pt}p{16pt}p{16pt}p{16pt}p{16pt}p{17pt}p{30pt}p{16pt}}
\toprule
& \textbf{Rule}          & \textbf{TN} & \textbf{FP} & \textbf{FN} & \textbf{TP} & \textbf{F-S} & \textbf{MCC/$w_e$} & \textbf{AUC} && \textbf{Rule}          & \textbf{TN} & \textbf{FP} & \textbf{FN} & \textbf{TP} & \textbf{F-S} & \textbf{MCC/$w_e$} & \textbf{AUC} \\
\midrule
\multicolumn{4}{l}{Client Characteristic}             &             &             &                  &               &              & \multicolumn{4}{l}{Need Satisfier}     &             &             &                  &               &              \\
& o-47$\sharp$ & 28          & 240         & 444         & 2605       & \textbf{0.88}$\S$            & -0.03         &              && o-44$\sharp$ & 2913        & 74          & 14996       & 2979        & 0.28             & 0.14          &              \\
& Aggr                   & 268         & 0           & 3049        & 0           & 0                & 0             & 0.52         & & s-46$\sharp$ & 2896        & 91          & 15042       & 2933        & 0.28             & 0.13          &              \\ \cline{1-9}
\multicolumn{4}{l}{Client Description}             &             &             &                  &               &              && o-36                   & 2701        & 286         & 13846       & 4129        & 0.37             & 0.11          &              \\
& p-34                   & 32          & 0           & 26          & 16          & 0.55             & 0.46          &              && o-2                    & 2916        & 71          & 16862       & 1113        & 0.12             & 0.06          &              \\
& o-42$\sharp$ & 31          & 1           & 42          & 0           & 0                & -0.13         &              && o-1                    & 2987        & 0           & 17950       & 25          & 0                & 0.01          &              \\
& o-43$\sharp$ & 31          & 1           & 42          & 0           & 0                & -0.13         &              && o-3                    & 2874        & 113         & 17301       & 674         & 0.07             & 0             &              \\
& Aggr                   & 32          & 0           & 26          & 16          & 0.55             & 0.46          & \textbf{0.70}$\flat$        && o-31                   & 2979        & 8           & 17931       & 44          & 0                & 0             &              \\ \cline{1-9}
\multicolumn{5}{l}{Desired State (outcome)}      &             &                  &               &              && o-16                   & 2962        & 25          & 17865       & 110         & 0.01             & -0.01         &              \\
& o-20                   & 1           & 44          & 3           & 40          & 0.63             & -0.11         &              && o-25                   & 2981        & 6           & 17958       & 17          & 0                & -0.01         &              \\
& Aggr                   & 45          & 0           & 43          & 0           & 0                & 0             & 0.52         && o-11                   & 2919        & 68          & 17847       & 128         & 0.01             & -0.06         &              \\ \cline{1-9}
\multicolumn{4}{l}{Need}             &             &    & &   &              & & s-12                   & 2856        & 131         & 17747       & 228         & 0.02             & -0.08         &              \\
& o-8                    & 36          & 7           & 69          & 6           & 0.14             & -0.13         &              && o-13                   & 1738        & 1249        & 13800       & 4175        & 0.36             & -0.15         &              \\
& o-9                    & 10          & 33          & 28          & 47          & 0.61             & -0.14         &              && o-22                   & 2628        & 359         & 17442       & 533         & 0.06             & -0.16         &              \\
& Aggr                   & 43          & 0           & 75          & 0           & 0                & 0             & 0.59         && Aggr                   & 2485        & 502         & 7994        & 9981        & 0.7              & 0.27          & \textbf{0.73$\flat$}        \\ \hline
\multicolumn{5}{l}{Need Satisfier Description}    &             &                  &               &              &                        &             &             &             &             &                  &               &              \\
& o-32                   & 5           & 0           & 62          & 21          & 0.4              & 0.14          &              &                        &             &             &             &             &                  &               &              \\
& o-18                   & 3           & 2           & 31          & 52          & \textbf{0.76}$\S$            & 0.11          &              & \multicolumn{4}{l}{Service Description}    &             &             &                  &               &              \\
& Aggr                   & 3           & 2           & 10          & 73          & 0.92             & 0.31          & \textbf{0.79$\flat$}        && o-23                   & 20          & 0           & 35          & 0           & 0                & 0             &              \\ \cline{1-9}
\multicolumn{4}{l}{Program Name} &             &             &                  &               &              && o-24                   & 20          & 0           & 35          & 0           & 0                & 0             &              \\
& s-35                   & 39          & 18          & 52          & 219         & \textbf{0.86}$\S$            & 0.42          &              && o-33                   & 20          & 0           & 35          & 0           & 0                & 0             &              \\
& s-44$\sharp$ & 55          & 2           & 209         & 62          & 0.37             & 0.19          &              && o-45$\sharp$ & 20          & 0           & 35          & 0           & 0                & 0             &              \\
& s-45$\sharp$ & 24          & 33          & 247         & 24          & 0.15             & -0.49         &              && o-15                   & 13          & 7           & 25          & 10          & 0.38             & -0.07         &              \\
& Aggr                   & 39          & 18          & 52          & 219         & 0.86             & 0.42          & \textbf{0.79$\flat$}        && s-41$\sharp$ & 19          & 1           & 35          & 0           & 0                & -0.18         &              \\ \cline{1-9}
\multicolumn{4}{l}{Required Criteria} &             &             &                  &               &              && s-43$\sharp$ & 19          & 1           & 35          & 0           & 0                & -0.18         &              \\
& o-28                   & 3           & 33          & 41          & 98          & \textbf{0.73}$\S$            & -0.2          &                                   && o-19                   & 14          & 6           & 30          & 5           & 0.22             & -0.19         &              \\
& Aggr                   & 36          & 0           & 139         & 0           & 0                & 0             & 0.61                              && Aggr                   & 20          & 0           & 35          & 0           & 0                & 0             & 0.66        \\
\bottomrule
\multicolumn{18}{l}{\footnotesize{$\sharp$ a rule based on T-chains  \,\,\, $\S$ a high F-score > $0.7$ \,\,\, $\flat$acceptable $AUC \geq 0.7$ value for a given $w_e$ model score.}}
\end{tabular}
\end{table*}

\section{Conclusion and Future Work}
In this paper, we propose a model for extracting SPO-related entities from descriptions. We also present the challenges and state of the NLP field, namely its lack of SPO-related corpora and pre-trained language models. Our model relies on our previous work for representing social services entities and semantics, namely the Common Impact Data Standard ontology~\cite{CIDS2021}, needed to capture an SPO's impact model as a ``narrative'' about their organization, services, and clients. Based on available data, the model relies on linguistic properties as well as empirically derived rules to identify phrases, construct $T$-triples, and classify phrases as entity types. Without any external data sources to seed the model with annotated text, the model performs well on certain entities, namely need satisfiers, program names, service descriptions, and required criteria. 

In future work, the model will be extended with additional features and training data. Negated phrases will generate semantics that negate a relationship, such as ``does not offer''. A larger corpus with correct annotations will allow for better scoring methods, more rules, the use of statistical methods, and the training of supervised machine learning models.  Finally, by incorporating available data associated with specific entities and extracted SPO narratives, we could perform analysis on an SPO's performance and suitability at a given time. For example, we cloud track a client's development as they transition from one program to another, based on the paths they take, the need satisfiers they qualify for, and ultimately use.


\bibliography{Gajderowicz-Rosu-Fox-text2story2022}

\end{document}